\documentclass[conference]{IEEEtran}
\usepackage[utf8]{inputenc}

\title{Pseudorehearsal in actor-critic agents}
\author{\IEEEauthorblockN{Marochko Vladimir\IEEEauthorrefmark{1},
Leonard Johard\IEEEauthorrefmark{2},Manuel Mazzara\IEEEauthorrefmark{3} }
\IEEEauthorblockA{Innopolis University
Email: \IEEEauthorrefmark{1}v.marochko@innopolis.ru,
\IEEEauthorrefmark{2}l.johard@innopolis.ru,
\IEEEauthorrefmark{3}m.mazzara@innopolis.ru
}}

\date{April 2017}
 \begin{document}

 \maketitle
 
\begin{abstract}
Catastrophic forgetting has a serious impact in reinforcement
learning, as the data distribution is generally sparse and non-stationary over time. The purpose of this study is to investigate whether pseudorehearsal can increase performance of an actor-critic agent with neural-network based policy selection and function approximation in a pole balancing task and compare different pseudorehearsal approaches. We expect that pseudorehearsal assists learning even in such very simple problems, given proper initialization of the rehearsal parameters.
\end{abstract}

\section{Introduction}

Reinforcement learning is a promising and growing area of machine learning based on training agents on positive or negative feedback to their actions. The agent gets an observation of the environment, chooses which action to perform, receives a reward and modifies its way of choosing future actions according to this reward. If the agent reached a state that is clearly better then the previous one in terms of the given task or successfully completed the task, then the reward is positive. If the agent reached a state clearly worse or failed the task, then the reward is negative.\\
Reinforcement learning algorithms can be divided into two different classes based on their ways to choose long-term optimal actions. One is based on generation of a model of the environment and evaluation of the states in this model to predict future reward. This is the class of model-based algorithms. The other - model-free algorithms - are based on trial-and-error learning based on habitual and conditioned responses tied to certain stimuli. We will focus on model-based algorithms.\\
The best situation is if the environment is fully observable and the number of states defined by agent is finite and not too big. In this case agent can just keep all possible states in the memory and evaluate each state directly. Of course, in real life this situation is very rare, so different kinds of approximation are used. One of the possible model approximations widely used is neural networks, which constitutes a powerful tool for implementation of memory. Unfortunately neural networks are vulnerable to a problem known as catastrophic forgetting. Within reinforcement learning influence of catastrophic forgetting is usually even more serious then in supervised learning and pseudorehearsal is one of possible ways of catastrophic forgetting elimination. We have shown that in Q-learning algorithms pseudorehearsal can improve performance significantly. \cite{marochko2017pseudorehearsal} and now want to test it on more interesting and complex actor-critic algorithm.\\
Actor-critic methods are one of the types of reinforcement learning model-based algorithms based on TD-learning. Actor-critic agents have a separate memory structure to explicitly represent the policy independent of the value function. Policy function is used to choose the next action. A value function evaluates the state reached by agent. Actor-critic agent takes an action chosen by the policy structure - actor. This actor is a probabilistic function from state to action. After that the agent receives reward and the critic consequently evaluates the action done. Finally, the critic's evaluation is used to update the actor. As a critic has two interacting neural networks, an actor-critic's vulnerability to catastrophic forgetting may be very serious.\\

\section{Theoretical issues}

\subsection{Actor-critic algorithm}

Actor-critic approaches achieves a high performance because they require minimal
computation in order to select actions – if the policy is explicitly stored, no computations are needed for action selection. Actor-critics can also learn an explicitly stochastic policy which is very useful in continuous learning problems which are common for reinforcement learning. Actor-critic algorithms, as stated above, uses the actor to choose actions based on the current state. The actor is typically a policy gradient function. Policy gradient actors can take the shape of a neural network and constructs distribution of actions probabilities, och which one is then selected at each step.

Policy gradient methods has good convergence properties, so they can reach its optimum quickly, but it does not store information about the environment. This means its performance is limited. Critics learn about and estimates the policy which is currently being followed by the actor. It provides a critique which takes the form of a TD error, which drives all learning in both the actor and the critic. The critic is typically a state-value function, which means it keeps information about environment and can predict which action will lead a better state. On the other hand the agent needs to explore an environment for a longer time to reach its optimum.\cite{beitelspacher2006policy}  After each action selection critic determines whether things have gone better or worse than expected. If TD error is positive - tendency to select the last action done is a for the future. If TD error is negative – this tendency should be weakened. Both the actor and the critic can use neural networks as a function approximation, which allows the agent to execute tasks in continuous and partially observable environments.\\

\subsection{Catastrophic forgetting}

Catastrophic forgetting is a common problem in neural networks. The problem occurs when a neural network that has properly learnt to execute some tasks meets changing conditions or learns the new task. This neural network learns the new task without any problems, but the old information might be almost fully erased in the process. This might not be a serious problem in supervised learning tasks like digit recognition where network was trained ones and never retrained, but in reinforcement learning,  where networks learning repeats regularly, e.g. at each episode or even each time step - catastrophic forgetting can seriously damage performance. \cite{cahill2011catastrophic} During the online learning in continuous space things become even worse. In this case the networks rarely receives real feedback and most of its learning iterations are based on its own assumptions. Even minor noise after one thousand step can make network to forget everything it has learnt.

The cause of catastrophic forgetting is based on the mathematical nature of neural networks. In linear networks that occurs because neural networks base their prediction from input data on the vector orthogonality. Two different sets with low orthogonality make the same neurons return different outputs on similar inputs, so that the learning of one set erases knowledge about the other. In the non-linear case the situation is not so clear, but tends to be similar - there is significant catastrophic forgetting when the information is very distributed and highly overlapping between sets. \cite{moe2005catastophic}

\subsection{Pseudorehearsal}
One of the methods or solving catastrophic forgetting is pseudorehearsal. It uses a two-step process: the first step is construction of the set of pseudopatterns and the second is training the network on pseudopatterns combined in batches with real patterns. The common way of pseudopattern construction is creating a set of pseudovectors, feeding them through the network and saving the outputs at each layer. This approach saves memory compared to rehearsal-based approaches because no real examples needed to be kept for catastrophic forgetting elimination. This approach doesn't involve any changes to the network structure as some other approaches do.\\
The generated approximations of the real data are sufficiently accurate in practice to reduce forgetting. Even extremely crude generative models have proven highly effective. In the original work in this area by \cite{robins1995catastrophic}, pure noise fed to the network was able to almost completely eliminate catastrophic interference. The argument of the authors was that, although the input is completely random, the activation distributions in deeper levels of the network will be representative of the learnt input data.\\
Pseudorehearsal methods have been demonstrated to significantly decrease and almost completely eliminate the catastrophic forgetting in unsupervised learning \cite{robins1995catastrophic}, supervised learning \cite{ratcliff1990connectionist} and reinforcement learning \cite{baddeley2008reinforcement}. It is interesting to note that the results of Baddeley suggest that the widely studied ill conditioning might not be the main bottleneck of reinforcement learning after all. Instead, their results indicate that the catastrophic forgetting is the main bottleneck for reinforcement learning problems.\\
We will try two basic pseudorehearsal approaches during our research. The one is using pseudopatterns for the correction of learning weights in learning with respect to orthogonality between the learned example and pseudovectors. The other is learning the network in classic batch-backpropagation way with one real example in a batch and others are pseudopatterns. We will vary size of the pseudopatterns batch and frequency of their reinitialization to find the ones which improve performance the most.

\section{Experimental design}

We apply pseudorehearsal algorithms to an actor-critic agent executing real reinforcement learning task. The task is a double pole-balancing problem, well-known reinforcement learning task mentioned for example by Sutton. \cite{sutton1998reinforcement}. The task of agent is to balance a two poles installed on a cart for as long as it possible by pushing the cart left or right. There is no positive reward in the problem and negative reward is given if any of poles falls down or the cart reaches the end of track. The primary result of these experiment is the number of steps for each agent can balance the pole. The bigger this number the better. Another secondary output is a time spent for computations. We want to know how good was the performance improvement compared to computational cost.

\subsubsection{Observation}
Two different observations are used for experimental comparison.
The first observation type given to the agents constitute a fully observable Markov decision process - the agent knows all the information about the current cart's state: cart position, velocity and acceleration and each poles' falling angle, angular velocity and angular acceleration. Therefore the best possible behaviour can be easily calculated through analytical methods, while the complex structures as neural networks create a huge amount of data redundancy which leads to a high amounts of catastrophic forgetting. For this reason this observation type is a very good one for testing tools of elimination of catastrophic forgetting. \\
The other observation type is a partially observable Markov decision process. In this case agent receives only information about current position of the cart and the poles. The neural networks are used appropriately - modelling the unknown environment. With this type of observation we might see how good is catastrophic forgetting in more realistic settings.\\
The observation is represented as a real valued vector, where each $i^{th}$ observed parameter is written into one of two vector cells: $2*i^{th}$ if the parameter is positive or $(2*i+1)^{th}$ if parameter is negative. The second vector entity assigned to parameter is assigned to zero. After that, the linear parameters are divided by 20 and angular are divided by 60 for normalization\\. 
\subsubsection{Performance metric}
Agent tries to complete the task for 1000 runs, then the number of successful steps and computation time in each try is saved. After that we can use different straight comparisons and statistical approaches to compare the results.The first step of straight comparisons is just subtraction of the one resulting vector from another, let's call the resulting vector difference vector. The next simple comparison practices are: drawing episode/step graphs for visual evaluation of agent's behaviour; drawing similar graphs but with mean of 10-20 consecutive steps instead of real values for evaluation of agent's tendencies, drawing these graphs for difference vectors. Of course these are just preparation of data for further statistical processing.\\
The main measures we use to analyze resulting vectors are the mean of the all resulting vector entities, the median and the root mean squared deviation. 

The mean of the vector tells us how good was the performance overall, the higher value the better. Root mean squared deviation tells us how strong was the influence of catastrophic forgetting, because cases of catastrophic forgetting intervention can be noted by significant change of performance between steps - usually to the side of decreasing. 

The median denotes how good is agent at removing bad information. Comparison of mean and median can give much information about the agent, which we can deduce from the nature of this parameters. If the median is higher than the mean, then this means that the largest deviations are on the side of lower than average performance. Therefore, the agent learns successfully and the catastrophic forgetting occasionaly has a strong impact, but this influence is then quickly eliminated. If the mean is higher then median - then while the agent tries to choose optimal policy, sometimes successfully, most of its runs are influenced by catastrophic forgetting being too high. If the mean and median are nearly the same then learning and forgetting counterbalance each other. Of course the higher absolute values of mean and median denote the higher overall performance of the agent and very important too. However, in some reinforcement learning tasks where the cost of mistake is high, like drone flight, we might choose the approach with worse performance, but with better catastrophic forgetting elimination, because reaching the goal slower is better then fast breaking of the machine in process. After the application of all this methods we need to apply significance test - like student's t-test to the results to see if our results are significant and therefore if we can make statement based on research.
\section{Conclusion}
This work will show us if the actor-critic algorithm is vulnerable to the catastrophic forgetting, and how good is pseudoreharsal in decreasing of this problem in the case of two interacting neural networks. We will find the best possible parameters of pseudorehearsal approaches and try to find dependency between the pseudorehearsal parameters and agent's performance and to explain this dependency.\\
If the pseudorehearsal well be proved to successfully eliminate catastrophic forgetting for continuous actor-critic algorithms, many reinforcement learning tasks will become easier to solve, and let agents quickly react on the instant changes in the environment. So this research may widen the sphere of the applications of reinforcement learning agents.
\bibliographystyle{ieeetr}
\bibliography{mybibliography.bib}
\end{document}